\pdfoutput=1
\documentclass[11pt]{article}

\usepackage[]{ACL2023}

\usepackage{times}
\usepackage{latexsym}

\usepackage[T1]{fontenc}
\usepackage[utf8]{inputenc}

\usepackage{microtype}

\usepackage{inconsolata}

\usepackage{soul}
\usepackage{url}
\usepackage{amsmath}
\usepackage{booktabs}
\usepackage{hyperref}
\usepackage{xcolor}
\usepackage{rotating}
\usepackage{multirow}
\usepackage{makecell}
\usepackage{graphicx}

\usepackage{xcolor,pifont}
\newcommand*\colourcheck[1]{%
  \expandafter\newcommand\csname #1check\endcsname{\textcolor{#1}{\ding{51}}}%
}
\newcommand*\colouruncheck[1]{%
  \expandafter\newcommand\csname #1uncheck\endcsname{\textcolor{#1}{\ding{53}}}%
}
\colourcheck{green}
\colouruncheck{blue}

\newcommand{\rewrite}[1]{\textcolor{black}{#1}}

\title{A Survey on Measuring and Mitigating Reasoning Shortcuts in Machine Reading Comprehension}

\author{
Xanh Ho,$^{\thefootnote{*} \; 1, 2}$
Johannes Mario Meissner,$^{\thefootnote{*} \; 3}$
Saku Sugawara,$^2$\and
Akiko Aizawa$^{1,2,3}$ \\
$^1$The Graduate University for Advanced Studies, Kanagawa, Japan\\
$^2$National Institute of Informatics, Tokyo, Japan \\
$^3$The University of Tokyo, Tokyo, Japan \\
{\tt \{xanh, saku, aizawa\}@nii.ac.jp} \\
{\tt jmariomeissner@gmail.com} 
}

\begin{document}
\maketitle

\begingroup\def\thefootnote{*}\footnotetext{Equal contribution.}\endgroup

\begin{abstract}
The issue of shortcut learning is widely known in NLP and has been an important research focus in recent years. Unintended correlations in the data enable models to easily solve tasks that were meant to exhibit advanced language understanding and reasoning capabilities. In this survey paper, we focus on the field of machine reading comprehension (MRC), an important task for showcasing high-level language understanding that also suffers from a range of shortcuts. We summarize the available techniques for measuring and mitigating shortcuts and conclude with suggestions for further progress in shortcut research. 
Importantly, we highlight two concerns for shortcut mitigation in MRC: (1) the lack of public challenge sets, a necessary component for effective and reusable evaluation, and (2) the lack of certain mitigation techniques that are prominent in other areas.\footnote{Resources are available at \url{https://github.com/Alab-NII/ReasoningShortcutsInMRC}.}
\end{abstract}

\section{Introduction}

% Introduction to MRC
Machine reading comprehension (MRC) is a task that requires the model to answer a given question by using the provided paragraphs.
To answer correctly, models need to connect and extract information across one or multiple paragraphs of text, exhibiting reading and reasoning skills; therefore, 
the MRC task is considered important for evaluating natural language understanding (NLU).
Many large-scale MRC datasets have been proposed, such as SQuAD\footnote{We denote SQuAD 1.1 as SQuAD.}~\cite{rajpurkar-etal-2016-squad}, RACE~\cite{lai-etal-2017-race}, HotpotQA~\cite{yang-etal-2018-hotpotqa}, DROP~\cite{dua-etal-2019-drop},  MuSiQue~\cite{trivedi-etal-2022-musique2}, and STREET~\cite{ribeiro2023street}.

\begin{figure}[htp]
\centering
    \includegraphics[scale=0.6501]{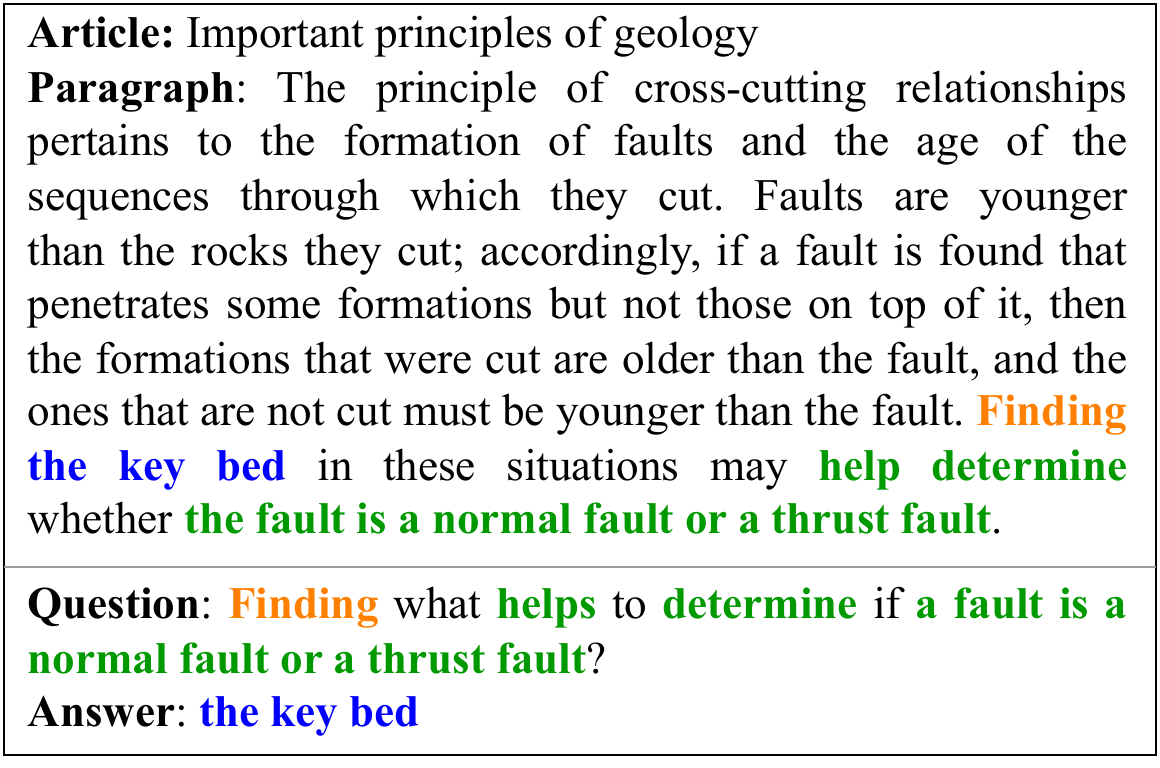}
    \caption{Example of reasoning shortcuts from SQuAD. The question can be answered by using word-matching (green) or the first word in the question (orange).}
    \label{fig:example}
\end{figure}
\renewcommand{\floatpagefraction}{.9}

% Leaderboard issue
Transformer-based models~\cite[\textit{inter alia}]{devlin-etal-2019-bert,Liu2019RoBERTaAR,NEURIPS2019_dc6a7e65,Clark2020ELECTRA,Lan2020ALBERT} have defeated humans on the SQuAD\footnote{\url{https://rajpurkar.github.io/SQuAD-explorer/}}
leaderboard. 
Although these results are impressive, these models are `brittle' when they are evaluated on adversarial examples or out-of-distribution (OOD) test data.
For example, via adversarial evaluation, \citet{jia-liang-2017-adversarial} demonstrate that current models do not understand natural language precisely.
\citet{sugawara-etal-2018-makes} show that many datasets contain a large number of `easy' questions that can be answered by only looking at the first few words of the question.
Figure~\ref{fig:example} presents an example from SQuAD, where the model can answer the question by using word-matching or the first word in the question.

% \cite[\textit{inter alia}]{app9183698,dzendzik-etal-2021-english,baradaran_ghiasi_amirkhani_2022,ijcai2022p0779,rogers2021qa}.

\begin{figure*}[tp]
    \includegraphics[width=\linewidth]{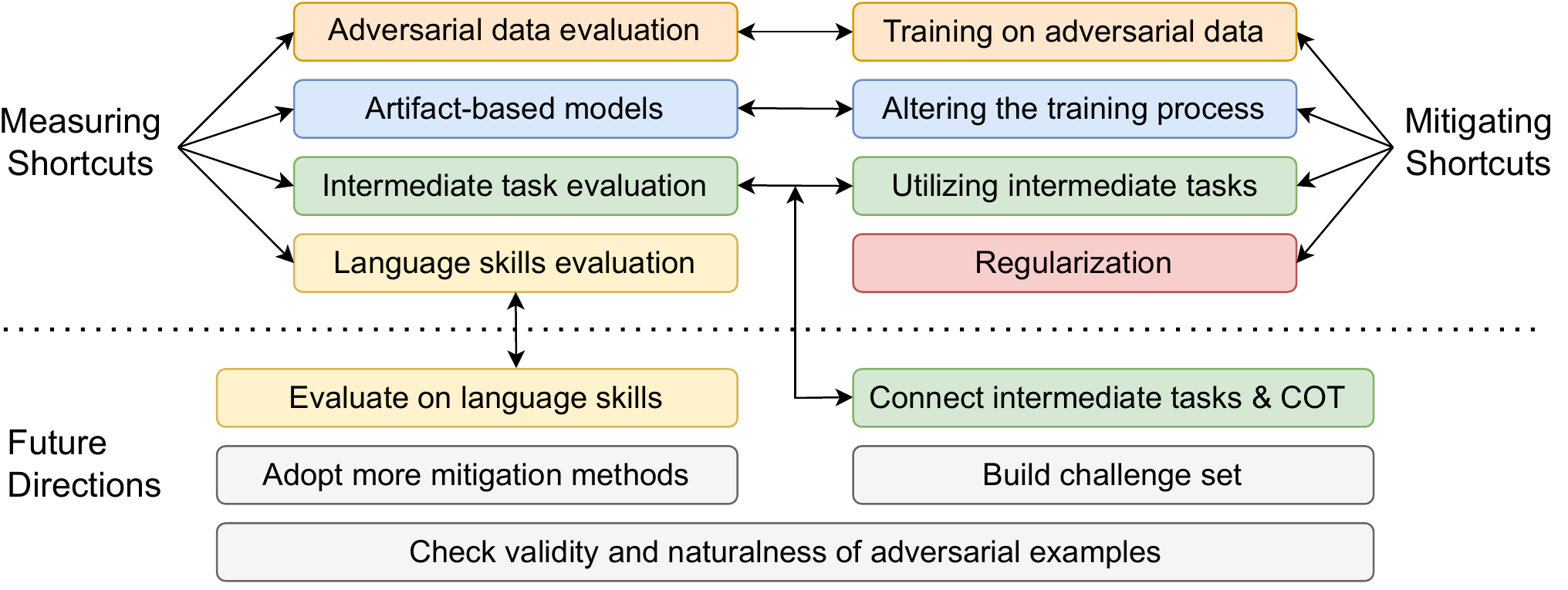}
    \caption{The techniques for measuring (Section \ref{sec_measure}) and mitigating (Section \ref{sec_prevent}) shortcuts we survey in this paper, as well as proposed future directions (Section \ref{sec_direction}).
    % COT denotes chain-of-thought.
    }
    \label{fig:overall}
\end{figure*}

Inspired by these findings, lots of studies have been proposed to detect, measure, and reduce reasoning shortcuts in MRC. 
This raises the need for a survey paper.
% existing surveys
There are many existing survey papers for MRC, 
such as papers that focus on datasets \cite{dzendzik-etal-2021-english,rogers2021qa} or systems/methods \cite{app9183698,baradaran_ghiasi_amirkhani_2022}.
To our knowledge, 
there is no existing survey paper specifically dedicated to reasoning shortcuts in MRC.
Although there are some survey papers on reasoning shortcuts \cite{Geirhos2020a,schlegel2020leaderboards,wang-etal-2022-measure,du2023shortcut},
they predominantly focus on general tasks in NLP or in larger domains such as computer vision. 
For example, \citet{Geirhos2020a} focus on shortcut learning in deep neural networks, including computer vision, while \citet{schlegel2020leaderboards}, \citet{wang-etal-2022-measure}, and \citet{du2023shortcut} 
cover a broader range of tasks in NLP, such as machine translation and natural language inference (NLI).

% In this paper
Different from existing survey papers, we go deeper into the MRC task, and highlight the shortcut detection and mitigation shortcomings that are unique to this task.
Specifically, we try to summarize and classify most existing studies to provide a broad-picture view for researchers on measuring and mitigating shortcuts in MRC (Section~\ref{sec_measure} and Section~\ref{sec_prevent}).
We also discuss several directions for future work (Section~\ref{sec_direction}).
Importantly, we highlight two main concerns for shortcut mitigation in MRC: the lack of public challenge sets and the lack of certain mitigation techniques that are prominent in other areas.
Figure~\ref{fig:overall} summarizes the techniques for measuring and mitigating shortcuts we survey in this paper, as well as proposed future directions.

% This paper is structured as follows. Section~\ref{sec_background} briefly introduces the background about MRC and relevant terminology. Section~\ref{sec_measure} and Section~\ref{sec_prevent} present the major shortcut detection and mitigation techniques respectively. 
% Then, Section~\ref{sec_direction} discusses directions for future work. Finally, the conclusion is summarized in Section~\ref{sec_conclusion}.
%
% It is noted that
% we do not offer an exhaustive citation list. Rather, we summarize and classify the most prominent studies across different approaches and methodologies.
%

\section{Background}
\label{sec_background}

\subsection{Machine Reading Comprehension Task}

% MRC is a sub-field within NLU where the text prompts given to the model are longer in nature, thus requiring reading comprehension capabilities to be solved.

MRC is a sub-field within NLU where the text prompts given to the model are longer in nature, thus requiring the model's reading comprehension capability.
Based on the answer format, four types of MRC datasets are available: span extraction, multiple-choice (MC), cloze style, and free-form answer~\cite{chen2018neural}.
In addition, to evaluate the multi-step reasoning ability of the models across paragraphs, \citet{welbl-etal-2018-constructing} introduce the multi-hop MRC task.
It requires a model to answer a given question by performing reasoning over multiple paragraphs.
% Option 1
Recently, conversational MRC tasks such as \citet{choi-etal-2018-quac} and \citet{reddy-etal-2019-coqa} have also been introduced.
For a comprehensive list of available datasets, we refer the reader to \citet{dzendzik-etal-2021-english}, \citet{sugawara-etal-2021-benchmarking}, \citet{yang2022more}, and \citet{rogers2021qa}.

\subsection{Definitions and Terminologies}

\paragraph{Reasoning Shortcut}
We define shortcuts as statistical correlations in the data that allow a machine learning model to achieve high performance on a task without acquiring all the intended knowledge. When these shortcuts happen in a task that was supposed to require a reasoning step, we denominate it reasoning shortcut.
The most important side-effect of shortcut learning is under-performance on adversarial or OOD data.

\paragraph{Adversarial Example} 
% Following \citet{schlegel2020leaderboards}, we define adversarial examples as those that are designed to mislead machine learning models but not humans. Usually, the perceived difficulty for a human remains unchanged, while models fail due to their shortcut behavior.

Following previous studies \cite{Geirhos2020a,schlegel2020leaderboards,10.1145/3374217}, we define adversarial examples as those that are designed to mislead machine learning models but not humans. Usually, the perceived difficulty for a human remains unchanged, while models fail due to their shortcut behavior.

% \paragraph{Independent and Identically Distributed (IID)}
% Following \citet{Geirhos2020a}, we consider data with the same statistical data distribution as the training set to be IID. 
% % Validation data that was randomly split from the training set is also IID.

% \paragraph{Out of Distribution (OOD)}
% Data with a statistical distribution different from the training data.

\paragraph{Challenge Set}
An evaluation dataset that highlights a particularly difficult aspect of a task, such as overcoming a prominent shortcut. These datasets are important to allow comparison between methods, and assess the progress made in shortcut mitigations. 
% In MRC, one of the most common challenge sets is Adversarial SQuAD, as introduced by \citet{jia-liang-2017-adversarial}.

\paragraph{Robustness and Generalization}
We define a model as robust if its performance remains relatively unaltered under adversarial attacks. Similarly, a model has the ability to generalize if it can perform well on OOD test data.

\section{Measuring Shortcuts}
\label{sec_measure}
% or detecting
Measuring the presence of shortcuts is an important first step necessary to understand the behavior of models and the biases present in the training data.
In this work, we divide the existing methods into four main groups from Sections \ref{evaluate-adversarial} to \ref{language_skill}.

% \subsection{Evaluate on Generalization Test}
\subsection{Adversarial Data Evaluation}
\label{evaluate-adversarial}

Adversarial samples are a clear and convenient way to highlight shortcut behavior because they are easy to construct and pose no additional difficulty for humans to solve.
Adversarial methods either add or edit an original example to create a more challenging example. Based on the change of the gold label, we divide this method into two main groups: (1) label-preserved (the answer is unchanged) and (2) label-changed (the answer is changed).
We summarize the methods we cover in Table~\ref{tab:measure_adversarial}.

% We observe that many methods can create grammatical examples; however, these examples are unnatural and real in the real world because the adding distractor sentence or distractor paragraph is not factual.

\begin{table*}[htp]
  \centering
    \resizebox{\textwidth}{!}{%
    \begin{tabular}{c l l l l c }
     \toprule
     % \textbf{Semantic} 
     & \multicolumn{1}{c}{\textbf{Name}}  & \makecell[c]{\textbf{Add/Edit}  \\ \textbf{Level} } & \makecell[c]{\textbf{Creation}  \\ \textbf{Method} } & \multicolumn{1}{l}{\textbf{Original Dataset}} & \multicolumn{1}{c}{\textbf{Naturalness}}    \\
    \midrule
    
    {\multirow{11}{*}{\rotatebox[origin=c]{90}{Label-Preserved}}}
    & AddSent~\cite{jia-liang-2017-adversarial} & sentence & semi-auto & SQuAD & \blueuncheck  \\
    
    % \makecell{AddSentDiverse \\ \cite{wang-bansal-2018-robust} }
   & AddSentDiverse~\cite{wang-bansal-2018-robust} & sentence & semi-auto & SQuAD &\blueuncheck \\

    & Negation~\cite{tran-etal-2023-impacts} & sentence & auto & SQuAD 2.0 & \blueuncheck \\

    & Universal~\cite{wallace-etal-2019-universal} & sentence & auto & SQuAD & \blueuncheck \\
   
   % \makecell{AddDoc \\ \cite{jiang-bansal-2019-avoiding} } 
    & AddDoc~\cite{jiang-bansal-2019-avoiding}  & paragraph & auto & HotpotQA & \blueuncheck  \\

    % & AddSent2Doc~\cite{ho-etal-2023-analyzing} & paragraph & auto & HotpotQA$^*$, 2Wiki & \blueuncheck \\

    % & Para-Perturb\todo{Rename}~\cite{wu-etal-2021-evaluating} & char/word/sent  & auto & SQuAD, TriviaQA & \blueuncheck \\  
    % char/word/sent various
    
    \cmidrule{2-6}
    
    & SEARs~\cite{ribeiro-etal-2018-semantically} & question & auto & SQuAD & \greencheck \\
    
    & Word-Replace~\cite{rychalska-etal-2018-are} & question & semi-auto & SQuAD & \greencheck  \\
    % \makecell{Ques-Paraphrase \\ \cite{gan-ng-2019-improving} }
    & Ques-Paraphrase~\cite{gan-ng-2019-improving} & question & semi-auto & SQuAD & \greencheck \\

    \cmidrule{2-6}
    % \makecell{Modify-Option \\ \cite{lin-etal-2021-using} }
    & Modify-Option~\cite{lin-etal-2021-using} & option & auto & RACE & \greencheck \\       
    
    \cmidrule{2-6}
    
    & Mix-Attack~\cite{si-etal-2021-benchmarking} & char/sent & auto & RACE & \blueuncheck \\

    & \cite{al-negheimish-etal-2021-numerical} &ques/pass & auto & DROP$^*$ & \blueuncheck \\   
    
    %  & aa~\cite{ribeiro-etal-2019-red} & level & & &  \\ 
    % \cmidrule{2-6}
    
    % & aa~\cite{} & level & & & \\
    
    % & aa~\cite{} & level & & &  \\ 
    
    \midrule
            
    {\multirow{5}{*}{\rotatebox[origin=c]{90}{Changed}}}

& Consistency \cite{ribeiro-etal-2019-red} & question & auto & SQuAD & \greencheck \\
    
    & Contrast Sets~\cite{gardner-etal-2020-evaluating} & word & experts & DROP, Quoref, ...  &  \blueuncheck \\

    \cmidrule{5-6}
    
    & SAM~\cite{schlegel-atal-2020-semantics} & word & auto  &   \makecell[l]{SQuAD, HotpotQA,  \\ DROP, NewsQA } & \blueuncheck  \\ 

     \cmidrule{5-6}

    & Break, Perturb, Build~\cite{geva-etal-2022-break} & question & auto & DROP, HotpotQA, IIRC & \greencheck \\      

     % & Unanswerable-Ques~\cite{nakanishi-etal-2018-answerable} & sentence & automated & SQuAD  & \greencheck \\

      \bottomrule
    \end{tabular}}
  \caption{Existing adversarial methods. 
  For each method, we present the name, the level at which the method performs modifications (e.g., word-level), the creation method, the original dataset, and the naturalness of the modified examples.
  Naturalness represents whether the modified example is natural and can occur in the real world.
  Dataset$^*$ represents a subset of the dataset that is used.
  }
  % or whether it can occur in the real world.
  \label{tab:measure_adversarial}
\end{table*}

\paragraph{Label-Preserved}

We divide this group into the following four types.

\textit{Context-modification}: \citet{jia-liang-2017-adversarial} is the first work that proposed adversarial examples for evaluating reading comprehension systems.
They insert distractor sentences into the original context of SQuAD (\texttt{AddSent}); their experimental results demonstrate that 
the current models fail to answer the modified examples.
% do not understand language precisely.
After that, \citet{wang-bansal-2018-robust} extend the \texttt{AddSent} method by adding the distractor sentence into various locations in the context (\texttt{AddSentDiverse}).
In addition to showing that the models are fragile on \texttt{AddSentDiverse}, they also showcase how their method improves robustness.
\rewrite{
In the same direction, \citet{tran-etal-2023-impacts} introduce a negation attack for SQuAD 2.0 to make models produce unanswerable responses to answerable questions.
They found that the performance of models trained on SQuAD 2.0 drops significantly on the negation attack.
} \citet{wallace-etal-2019-universal} propose a universal adversarial attack by using gradient-guided search for many tasks in NLP, including MRC.
Their triggers can attack 72\% of `why' questions in SQuAD, making them produce the same answers.
They also reveal that the models are based heavily on the words around the answer in the paragraph and question types when producing an answer.

For the multi-hop MRC task, instead of adding distractor sentences, \citet{jiang-bansal-2019-avoiding} add a distractor paragraph.
They demonstrate that many examples in the HotpotQA dataset contain reasoning shortcuts, where the models can answer the question by using word matching.
% 
% \rewrite{
% \citet{ho-etal-2023-analyzing} observe that HotpotQA and 2WikiMultiHopQA~\cite[2Wiki;][]{ho-etal-2020-constructing} contain position bias, where most of the sentences used to answer the questions are the first sentences in the paragraph.
% %
% They then create adverarial sets by adding distractor sentences at the beginning of each paragraph.
% The results show that the performance of the models drops on these adversarial sets. 
% }

% 
% \citet{wu-etal-2021-evaluating} performed various experiments by using three different levels of perturbations: character, word, and sentence.
% They also performed experiments with different embedding types and different amounts of the dataset.
% In general, their detailed experimental results revealed that the models are easy to attack.

\textit{Question-modification}: \citet{ribeiro-etal-2018-semantically} introduce a set of rules that modify some characters in the question but still keep the semantics. 
These are called semantically equivalent adversarial rules (SEARs).
Their experimental results show that models are weak to these changes; the model's predictions are changed after applying these rules.
After that, 
\citet{rychalska-etal-2018-are} modify the questions by changing some important words using the LIME (Locally Interpretable Model Agnostic Explanations) framework~\cite{lime}.
They show that performance decreases when some words in the questions are replaced with their synonyms.
Later in this direction, \citet{gan-ng-2019-improving} introduce two approaches to paraphrase the questions: (1) make them similar to the original questions to test model over-sensitivity, and (2) use context words near an incorrect answer candidate.
They show that their models drop in performance on both types of paraphrased questions.

\textit{Option-modification}:
\citet{lin-etal-2021-using} modify options in the multiple-choice dataset RACE while keeping the passage and the question.
Specifically, they replace one wrong option among the four candidates with an irrelevant option which is chosen from a set of options via experiments.
Their results reveal that the models exploited statistical biases in the datasets when answering the questions.
% the current models

\textit{Mix-modification}:
\citet{si-etal-2021-benchmarking} introduce four types of adversarial attacks and create a new benchmark for evaluating robustness in MRC.
Their dataset, AdvRACE, is created by modifying RACE.
%
% The four adversarial attacks are AddSent, CharSwap, Distractor extraction, and Distractor generation; these attacks include all three types of modifications: context, question, and option.
%
Their experimental results show that the models are vulnerable under all of these attacks; meanwhile, their dataset, AdvRACE can be served as test data for evaluating robustness.
\rewrite{
\citet{al-negheimish-etal-2021-numerical} evaluate top-performing models in the DROP leaderboard on a variety of modified versions of the DROP dataset.
% (a subset about numerical reasoning), including question modification and passage modification.
%
They find that the models do not reason about the question and the content of the passage, instead exploiting spurious patterns in the dataset to obtain the answers.
}

\paragraph{Label-Changed}
\rewrite{
For the case when the dataset modifications include a change in the answer, \citet{ribeiro-etal-2019-red} automatically generate new question-answering (QA) pairs that represent the same information as the original QA but in a different way. 
% For example, the original QA is: Q: When did Zhenjin die? A: 1285; the new QA is: Q: Who died in 1285? A: Kublai (wrong prediction).
Via evaluation, they find that 
the models lack real comprehension skills for their correct predicted answers in the original QA samples. 
}
\citet{gardner-etal-2020-evaluating} manually create contrast sets for 10 NLP datasets (including MRC) where the context is slightly modified and the answer is changed.
Their experimental results show that model performance dramatically drops on the contrast sets.
\citet{schlegel-atal-2020-semantics} create a SAM (Semantics Altering Modifications) dataset by modifying the context that makes a change in the answer.
They reveal that most of the current models are struggling with their proposed dataset.
\rewrite{
More recently,~\citet{geva-etal-2022-break} automatically generate adversarial samples by changing the reasoning path through question decomposition.
% Specifically, they first (1) decompose a question into multiple reasoning steps, then (2) change these reasoning steps, and finally obtain new questions based on the new reasoning steps.
% Their results reveal that the performance of the models drops on their adversarial dataset.
%
Their results reveal that the performance of the models drops on the adversarial dataset
}
Additionally, several studies~\cite{nakanishi-etal-2018-answerable,rajpurkar-etal-2018-know,trivedi-etal-2020-multihop,trivedi-etal-2022-musique2} introduce unanswerable questions by modifying the context or adding new questions.
% where the answers were changed

\paragraph{Adversarial Datasets}
From the above-mentioned methods, we have several adversarial or challenge datasets, such 
as Adversarial SQuAD and AdvRACE.
% as Adversarial SQuAD~\cite{jia-liang-2017-adversarial}, SAM~\cite{schlegel-atal-2020-semantics}, and AdvRACE~\cite{si-etal-2021-benchmarking}.
%
Additionally, some others are created by humans or a human-and-model in the loop process.
Specifically, \citet{khashabi-etal-2020-bang} apply human-driven natural perturbations to create a natural perturbation dataset (Natural-Perturbed-QA) in which the answer can be changed or unchanged when compared with the original example.
%
% \citet{bartolo-etal-2020-beat} and \citet{kiela-etal-2021-dynabench} applied human-and-model in the loop to create the two datasets, AdversarialQA and Dynabenchmarking, respectively.
\citet{bartolo-etal-2020-beat} use a human-and-model-in-the-loop method to create the AdversarialQA dataset.
Similar to~\citet{bartolo-etal-2020-beat}, \citet{kiela-etal-2021-dynabench} introduce the Dynabench framework, which supports human-and-model-in-the-loop dataset creation.
They apply their framework for four NLP tasks, including the MRC task.
\rewrite{
Different from all the mentioned approaches, \citet{pmlr-v119-miller20a} introduce four new test sets for SQuAD, with three of them aiming to measure robustness to natural distribution shifts.
They find that models are less robust on two of these three datasets, while human results remain the same.
}
\rewrite{All adverserial datasets are summarized in Appendix~\ref{sec_measuring_appendix_adver}.}

\subsection{Artifact-Based Models} \label{artifact-models}
% To detect weaknesses and reasoning shortcuts, there are several artifact-based models have been proposed, such as the question-only model, passage-only model, and single paragraph-only model (for multi-hop MRC task).
% The idea behind these models is that the input data in which the models are trained on is insufficient to answer the question.
% If the models can perform well on these examples, we can infer that the models perform shortcuts learning in the QA process.

Artifact-based models are trained on insufficient or incomplete data, such as question-only, passage-only, or single-paragraph-only (in multi-hop tasks). 
If these models perform well, it can be inferred that the missing information was not necessary, and shortcuts were used within the provided data.

\citet{kaushik-lipton-2018-much} perform various experiments on 5 MRC datasets 
%
% (CBT~\cite{journals/corr/HillBCW15}, CNN~\cite{NIPS2015_afdec700}, Who-did-What~\cite{onishi-etal-2016-large}, bAbI~\cite{DBLP:journals/corr/WestonBCM15}, and SQuAD) 
%
by using two artifact-based models: question-only and passage-only.
Their results reveal that the models can achieve higher scores when they are trained in this way.
For example, in task 18 of the bAbI dataset, the question-only approach obtains 91\%, while the best performance of a standard model is 93\%.
These results indicate that the models are not solving the task in the manner expected, and instead abuse shortcuts.
After that, \citet{Si2019WhatDB} use the same methods as \citet{kaushik-lipton-2018-much}; they also add another experiment by shuffling the words in the context.
They suggest that there exist artifacts and statistical cues in five MC datasets.
% : RACE, MCTest~\cite{richardson-etal-2013-mctest}, MCScript~\cite{ostermann-etal-2018-mcscript}, MCScript2.0~\cite{ostermann-etal-2019-mcscript2}, and DREAM~\cite{sun-etal-2019-dream}.

In multi-hop MRC, where at least two paragraphs are required to answer the question, \citet{min-etal-2019-compositional} and \citet{chen-durrett-2019-understanding} design a sentence-factored model and a single-paragraph BERT-based model respectively.
The introduced models are not trained in the full context; therefore, they should not have the ability to answer the questions.
However, their results show that these models can answer a large portion of examples; this indicates that these models do not perform multi-hop reasoning in the QA process.
With the same idea, \citet{trivedi-etal-2020-multihop} introduce the DiRe (Disconnected Reasoning) condition by removing the connection of the two supporting facts to measure reasoning shortcuts.
They conclude that there had not been much progress in multi-hop reasoning.

Different from the above works, \citet{sugawara-etal-2018-makes} use the first few words in the question instead of using the full question.
It was revealed that the BiDAF model~\cite{seo2017bidirectional} can infer the answer by using entity type matching.
\citet{sen-saffari-2020-models} expand the idea in \citet{sugawara-etal-2018-makes} by using BERT and find 
% the same behavior that the models can answer the questions without using most or all of the words in the question.
that again, the model can answer the questions without using most or all of the words in the question.

There is one special case in this group where a shortcut is detected by using a subset of the dataset.
Specifically, \citet{ko-etal-2020-look} demonstrate the presence of the position bias in the SQuAD dataset by training the models on a subset of SQuAD in which the answer is in the first sentence of the context.
Model performance drops significantly when evaluated on the SQuAD development set.

\subsection{Intermediate Reasoning Task Evaluation}
\label{sec_detect_inter}

The underlying reasoning process from question to answer is important information to verify whether the models know how to answer the question in a step-by-step manner.
One special requirement for this method is that it is only applicable to complex questions, such as multi-hop questions.
The reasoning steps from question to answer of complex questions can be used to design new sub-tasks or to evaluate the reasoning abilities of the model.

\rewrite{
In general, there are two main types of questions in the multi-hop MRC task: bridge and comparison questions.
\citet{tang-etal-2021-multi} simply evaluate the underlying reasoning process via a set of sub-questions for bridge questions.
It reveals that the existing multi-hop models do not have the ability to answer the sub-questions well, and many of them are answered incorrectly while their corresponding multi-hop questions are correctly predicted.
After that,~\citet{ho-etal-2022-well} evaluate the underlying reasoning process for comparison questions by introducing the HieraDate dataset with three probing sub-questions: extraction, reasoning, and robustness.
They find that even when the model is fine-tuned on the reasoning sub-questions, it does not have the ability to subtract two dates, although it can subtract two numbers.
}

% NumNet+ \cite{ran-etal-2019-numnet}
% They show that without fine-tuning, the models does not have abilities to perform the reasoning tasks. However, when the models are fine-tuned on these probing sub-questions, the performance improve significantly.

\citet{inoue-etal-2020-r4c} and \citet{ho-etal-2020-constructing} propose a new task for predicting or generating the reasoning chain; it is called derivation prediction in $\mathrm{R^4C}$ and evidence generation in 2WikiMultiHopQA~\cite[2Wiki;][]{ho-etal-2020-constructing}. 
\citet{wolfson-etal-2020-break} propose the `Break It Down' dataset that contains an ordered list of steps in the process from question to answer. However, these steps are only used for training, not for evaluation.
\rewrite{
After that,~\citet{10.1162/tacl_a_00370} introduce the StrategyQA dataset with sub-questions to explain the answers.
% The unique characteristic of StrategyQA is that its questions are implicit, meaning that there is no explicit information in the questions themselves for the question decomposition process.
%
Recently,~\citet{trivedi-etal-2022-musique2} propose the MuSiQue dataset, which is constructed via single-hop question composition.
% One advantage of MuSiQue is that it contains the sub-questions for the QA process.
% However, unfortunately the authors do not include the sub-questions evaluation when comparing between models in their paper.
Most of these existing multi-hop datasets contain only a small number of reasoning steps, which are easy for the models.
To solve this issue,~\citet{ribeiro2023street} introduce the STREET dataset with more reasoning steps in the QA process.
STREET requires a model not only to predict an answer for the question but also to generate a step-by-step structured explanation to explain the answer.
Their results reveal that few-shot prompting GPT-3 and fine-tuned T5 do not possess sufficient skills to generate the structured reasoning steps.
%
% Before STREET, there is a dataset called EntailmentBank \cite{dalvi-etal-2021-explaining}, which also requires the models to generate a structured explanation (a tree form). However, EntailmentBank only focuses on one task, which is tree generation, and does not include the QA task.
}

% Unfortunately, these models do not control all steps in one model to detect how effective the intermediate tasks are.

\rewrite{
We summarize all the studies mentioned above in Appendix~\ref{sec_measuring_appendix_inter}.
As shown in Appendix~\ref{sec_measuring_appendix_inter},
currently, there are many forms (e.g., sub-questions and triples) used to represent the reasoning steps from the question to the answer. 
However, it is still not clear what the different effectiveness of each form is for measuring and mitigating shortcuts.
}

There are some other related studies to this approach, such as building more explainable models~\cite{min-etal-2019-multi,fu-etal-2021-decomposing-complex} by using question decomposition~\cite{perez-etal-2020-unsupervised}.

% \subsection{Language Skills Evaluation}
\subsection{Language Understanding Skills Evaluation}
\label{language_skill}
For humans to answer the question in the MRC task correctly, it requires several skills such as entity linking or coreference resolution. 
%
% As an intrinsic feature, the QA process in the MRC task also involves multiple skills.
%
In this section, we explore approaches that evaluate models on these basic skills. Models that answer the task correctly but fail at the required skills for the task can be said to be abusing shortcuts.

\citet{ribeiro-etal-2020-beyond} introduce CheckList, a list of basic linguistic skills to test the models comprehensively.
Includes several skills, such as temporal reasoning, negation, coreference resolution and semantic role labeling.
Their results show that models do not have the abilities to handle these skills.
As an example, given the context ``Aaron is an editor. Mark is an actor.'' and the question ``Who is not an actor?'', the model incorrectly predicts ``Mark''.
\rewrite{
At the same time, \citet{dunietz-etal-2020-test} introduce a template of understanding, which is ``a set of question templates'', to systematically test the comprehension abilities of the models regarding the content.
Through a pilot experiment, they show that the XLNet \cite{NEURIPS2019_dc6a7e65} model performs worse on their designed questions. 
}

\citet{wu-etal-2021-understanding} introduce seven MRC skills that are related to discourse relations, such as negative causality reasoning and explicit conditional reasoning.
Their results show that the three datasets (SQuAD, SQuAD 2.0, and SWAG~\cite{zellers-etal-2018-swag}) are insufficient for evaluating the understanding of discourse relations.
Prior to this, \citet{DBLP:conf/aaai/SugawaraSIA20} analyze 10 datasets with 12 requisite skills.
They also conclude that most existing MRC datasets might be insufficient for evaluating the discourse relations understanding.

We argue that evaluating the models on more basic NLP skills is an effective way to ensure that the models follow what humans do in the QA process. 
Future studies for carefully designing a set of language skills corresponding to each evaluation test data would be a need.

\subsection{Summary \& Discussion}

% The four introduced methods can be divided into two types: external and internal.
% Adversarial data evaluation and artifact-based models are the external methods; meanwhile, intermediate reasoning task evaluation and language understanding skills evaluation are the internal methods.
% %
% The external methods do not focus on the reasoning steps from question to answer. In contrast, the internal methods focus on evaluating the reasoning steps in the QA process or evaluating related basic language understanding skills that are necessary for humans in the QA process.
% %
% Both methods can detect the existence of shortcuts in the QA process of the current MRC models.
% It seems that the internal methods are more explainable and more similar to humans in the QA process than the external methods.

In addition to the studies belonging to these groups, there are several methods that detect reasoning shortcuts by manually analyzing the datasets~\cite{pugaliya-etal-2019-bend,schlegel-etal-2020-framework,lewis-etal-2021-question}.
\rewrite{
For example, via analysis, \citet{lewis-etal-2021-question} show that there is a significant overlap between the train and test sets.
Via experiments, they show that the model performance is worse on test examples that do not overlap with the training set.
}

The four introduced methods can be divided into two types: external and internal.
Adversarial data evaluation and artifact-based models are the external methods; meanwhile, intermediate reasoning task evaluation and language understanding skills evaluation are the internal methods.
The external methods do not focus on the reasoning steps from question to answer. In contrast, the internal methods focus on evaluating the reasoning steps in the QA process or evaluating related basic language understanding skills that are necessary for humans in the QA process.
Both methods can detect the existence of shortcuts in the QA process of the current MRC models.
It seems that the internal methods are more explainable and more similar to humans in the QA process than the external methods.

% chen-etal-2016-thorough,

% In summary, there are four types of shortcuts that the existing works have detected: entity type~\cite{sugawara-etal-2018-makes,min-etal-2019-compositional}, word overlap~\cite{jiang-bansal-2019-avoiding}, statistical bias~\cite{lin-etal-2021-using}, and position bias~\cite{ko-etal-2020-look,ho-etal-2023-analyzing}.
\rewrite{
In summary, there are four types of shortcuts that the existing studies have detected: \textit{entity type}, \textit{text overlap} (it includes question-passage overlap and train-test overlap), \textit{statistical bias}, and \textit{position bias}.
However, there may still be many other types of shortcuts that we have not discovered yet.
}

\section{Mitigating Shortcuts}
\label{sec_prevent}

% \rewrite{
% Shortcut mitigation techniques have been developed in most, if not all, fields of machine learning. However, a certain level of fragmentation exists between fields, and not all methods have been thoroughly explored everywhere. We mention some approaches that, to our knowledge, have not been applied to MRC yet, but are common in other areas of NLP, such as Natural Language Inference.
% }

Most shortcut measuring approaches detailed in the previous section can be leveraged for mitigation purposes. Thus, we follow a similar order: training on several kinds of adversarial data, followed by leveraging artifact models and data subsets, then utilizing the intermediate reasoning tasks, and finally a brief mention of other approaches.

\subsection{Training on Adversarial Data}

% \paragraph{Adversarially Created Datasets} 
\paragraph{Adversarial Datasets} 
Adversarial data was described in \ref{evaluate-adversarial} as a vital approach to highlight unwanted model behaviors. However, it can also be used as an effective mitigation approach. As a matter of fact, most studies that collect or generate adversarial data also use it as training data to analyze the robustness improvements. 
For example, both \texttt{AddSent} \cite{jia-liang-2017-adversarial} and \texttt{AddSentDiverse} \cite{wang-bansal-2018-robust} have trained models on their newly generated data. While \texttt{AddSent} showed only `limited utility', \texttt{AddSentDiverse} achieved `significant robustness improvements'. The more elaborate \texttt{AddDoc}, which appends whole documents as an additional source, shows `statistically significant improvements' when training on the generated data.

While the above methods use automatically generated data, it is also common to refer to human crowdsourcing to generate adversarial samples. 
\citet{bartolo-etal-2021-improving} propose to improve the quality of such adversarially collected data with automatic procedures, achieving even stronger results. Naturally, manual annotation usually yields more diverse data at a higher cost, but suffers from the well-known human annotation artifacts. Arguably, any adversarial dataset can be used for training, and should in a majority of cases contribute towards robustness. \citet{liu-etal-2019-inoculation} illustrate this clearly: including just a few samples of adversarial data in the training process greatly improves scores, a clear indication of their debiasing potential.

UnifiedQA \cite{khashabi-etal-2020-unifiedqa} was recently introduced as a new state-of-the-art model for question answering, its strong-point being that it was trained on a large variety of datasets at once. Yet, limitations on generalizability are still being observed. In \citet{le-berre-etal-2022-unsupervised}, an unsupervised question generation process is developed that, similarly to other adversarial data generation methods, contains distractor sentences. This additional data helps UnifiedQA improve its out-of-domain generalization. This research direction shows that, even at extreme data and model sizes, generalization remains an issue and adversarial data generation can continue to help improve performance.

\paragraph{Unanswerable Data} 
Unanswerable questions are samples in the dataset that require a new `unanswerable' label, and the model should correctly identify the questions that have no answer. The main goal is to avoid over-confidence, which contributes to the reduction of shortcut abuse.

% \citet{rajpurkar-etal-2018-know} create an augmented SQuAD dataset with unanswerable questions (SQuADRUn), and models are trained with the additional task of predicting the answerability. 
\citet{trivedi-etal-2020-multihop} introduce a method to transform datasets to include `disconnected reasoning' evaluation in HotpotQA. Essential paragraphs required for a complete reasoning chain are programmatically removed, and the new data can be used to train a model to predict when the question is answerable, which in turn reduces shortcut-learning behavior.
\citet{lee-etal-2021-robustifying} obtain pseudo-evidentiality labels for HotpotQA, such that the training set contained both questions with enough evidence and questions without. On the latter, the model is discouraged from achieving high confidence by modifying the training objective. 
\rewrite{
\citet{tran-etal-2023-impacts} compare models trained on answerable samples only (SQuAD) and models trained on both answerable and unanswerable samples (SQuAD 2.0).
% They initially find that models trained on SQuAD 2.0 do not improve the robustness of the models on adversarial attacks when compared with those trained on SQuAD.
% Through error analyses, they introduce a technique called `force to answer'.
% By using this technique, they show that models trained on SQuAD 2.0 are more robust than those trained on SQuAD for answerable questions.
% Additionally, models trained on SQuAD 2.0 also demonstrate greater robustness on additional out-of-domain datasets.
By introducing a technique called `force to answer', they show that models trained on SQuAD 2.0 are more robust than those trained on SQuAD for answerable questions.
Additionally, models trained on SQuAD 2.0 also demonstrate greater robustness on additional OOD datasets.
}
%
% Unfortunately, no comparisons are drawn against a model trained without it, so additional work is necessary to conclude if this method truly helps to avoid shortcuts.

\paragraph{Data Perturbations}
Data perturbations or augmentations have been standard practice in computer vision for a long time, and only significantly entered the NLP in more recent years. \citet{feng-etal-2021-survey} outline the main approaches that exist to date. As an important subfield, perturbations can be made to the embedding space instead of the input token space. \citet{liu2020adversarial} perform this technique on the SQuAD dataset, with this being the only notable mention for MRC.
Tangentially, while not an MRC application, we see potential in the approach taken by \citet{wang-etal-2022-identifying}, where individual biased words are masked out, leading to debiasing benefits.

\subsection{Altering the Training Process}

The approaches in this section utilize some additional information to either modify the loss function, or filter out / select certain samples in the training data.

\paragraph{Debiasing Losses}

For this technique, we require a purposefully biased model whose predictions will be used to estimate the level of bias in each sample. These models can be created as described in Section~\ref{artifact-models}. The bias level can be leveraged during train time, modifying the loss function to discourage bias learning. 
% In the next paragraph, we detail the main losses, and then we further mention some works applying this method to MRC tasks.
% The approaches can be roughly grouped into three strategies. 

Product-of-expert \cite{hinton-2002} combines the logits of the biased model with those of the target model. The target model is thus forced to take the lead in learning non-biased behavior, as the bias is already being provided. A successor to this strategy, Learned Mixin \cite{clark-etal-2019-dont}, has a learnable parameter controlling the balance between the logits of both models.
On the other hand, Sample Reweighting \cite{schuster-etal-2019-towards} reduces the contribution of biased samples towards the loss, while Confidence Regularization \cite{utama-etal-2020-mind} aims to encourage the model to produce higher-entropy answers for biased questions by smoothing the labels based on the bias level. 

In MRC, we have seen a few major applications of these methods. \citet{clark-etal-2019-dont} compare Product-of-Expert against Learned Mixin on the SQuAD dataset, while \citet{wu-etal-2020-improving} apply Confidence Regularization leveraging multiple biases simultaneously by incorporating the biased predictions of several models into the loss function. It is one of the more comprehensive debiasing works in MRC, with five datasets being used to evaluate the approach: SQuAD, HotpotQA, TriviaQA~\cite{joshi-etal-2017-triviaqa}, NewsQA~\cite{trischler-etal-2017-newsqa} and Natural Questions~\cite{10.1162/tacl_a_00276}.

The case of removing unknown shortcuts or biases is treated as a separate challenge. In this scenario, we cannot build artifact-based models; instead, we use the concept of weak models. \citet{sanh-etal-2021-learning} propose using under-parameterized models, under the assumption that these models learn mostly shortcuts. Their method is evaluated on SQuAD. Similarly, \citet{utama-etal-2020-towards} proposed utilizing an under-trained model. \citet{ghaddar-etal-2021-end} improve the technique even further by integrating the bias model (a simple attention-based classification layer) into the main model, and training both simultaneously. This brings the benefit of avoiding the separate weak model training stage. Generally speaking, there is a lack of work regarding the removal of unknown shortcuts in the field of MRC, and we expect this area to experience more attention in the coming years.

\paragraph{Valuable Subsets}
We include in this category methods that select a subset of the data for re-training or additional training, with the goal of improving robustness. We mention these methods due to their importance in other NLP tasks, but their application in MRC is yet to be seen.
The existing methods for this technique are presented in Appendix~\ref{sec_mitigating_appendix_valuable}.

\subsection{Utilizing Intermediate Reasoning Tasks}
\rewrite{
The idea behind this approach is to determine whether utilizing the intermediate tasks along the path from question to answer can improve the robustness of the models. 
Currently, there are two well-defined tasks in the QA process.
The first one is extractive rationale prediction~\cite{lei-etal-2016-rationalizing,chen-etal-2022-rationalization}. 
The extractive rationale is referred to as a highlight in \citet{wiegreffe-marasovic-2021-review}.
This information is also similar to the `sentence-level supporting facts (SFs)' in the multi-hop MRC task. 
This task is often formulated as a binary classification task, where the objective is to predict sentences or words that appear in the context and can be used to answer the questions.
The second task is reasoning chain prediction.
This task is similar to the task in~\citet{inoue-etal-2020-r4c} and~\citet{ho-etal-2020-constructing} (refer to Section~\ref{sec_detect_inter} for details).
}

\rewrite{
% Similar to~\citet{jia-liang-2017-adversarial},
%
% : AddText-Rand, AddText-Wiki, and AddText-Adv
\citet{chen-etal-2022-rationalization} propose three types of attacks and conduct experiments on 5 datasets, including two MRC tasks (SQuAD and MultiRC \cite{khashabi-etal-2018-looking}), to verify whether rationalization can improve the robustness of the models.
Their results reveal that explicitly training the models with the rationale prediction task does not guarantee the robustness of the models on attacks.
\citet{ho-etal-2023-analyzing} utilize both sentence-level SFs task and entity-level reasoning task in the multi-hop MRC task for their training.
%
% They find that the model trained with intermediate tasks can help prevent the position bias in the 2Wiki dataset but cannot prevent the position bias in a subset of HotpotQA.
They find that the model, trained with intermediate tasks, reduces position bias in the 2Wiki dataset but not in a subset of HotpotQA.
%
% They explain that this result is due to the different levels of position bias and the difference in the percentage of comparison and bridge questions between these two datasets.
%
They attribute this result to varying levels of position bias and differences in the percentage of comparison and bridge questions across the datasets.
In summary, these results are quite sensitive to the datasets. Currently, there are no comprehensive analyses that fully exploit the effectiveness of intermediate tasks to prevent reasoning shortcuts and biases.
}

\subsection{Other Approaches}

InfoBERT \cite{wang2021infobert} was recently proposed as a way to control the amount of information that is learned during the fine-tuning stage. The goal is to use the mutual-information measure to filter out noise and obtain more robust representations. Promising results have been obtained in several tasks including SQuAD.

\begin{table*}[ht]
\resizebox{\textwidth}{!}{%
\centering
\begin{tabular}{@{}lll@{}}
\toprule
\multicolumn{1}{c}{\textbf{Name}} & \multicolumn{1}{c}{\textbf{Type/Sub-type}}   & \multicolumn{1}{c}{\textbf{Evaluation Dataset}}   \\ \midrule
AddSentDiverse \cite{wang-bansal-2018-robust}
                     & \multirow{3}{*}{Adversarial} & Adversarial SQuAD  \\
AddDoc \cite{jiang-bansal-2019-avoiding}
                     &                              & Add4Docs, Add8Docs \\
Synthetic Adversarial Generation Pipeline \cite{bartolo-etal-2021-improving}
                     &                              & AdversarialQA      \\
                     
                     \cmidrule{2-3}
                     
Disconnected Reasoning \cite{trivedi-etal-2020-multihop}
                     &    \multirow{3}{*}{Unanswerable}     & Custom     \\
Pseudo-evidentiality in HotpotQA \cite{lee-etal-2021-robustifying}
                     &                              & Custom             \\
Force to Answer \cite{tran-etal-2023-impacts}
                     &                               &    SQuADRUn 
                     \\ 
                     
                      \cmidrule{2-3}

Embedding Space Perturbations \cite{liu2020adversarial}       
                     & Perturbations                & Adversarial SQuAD  \\ \midrule

Learned-Mixin \cite{clark-etal-2019-dont}     
                     &             \multirow{3}{*}{Bias Labels}                 & Non-Adv OOD Tasks  \\
                     
Multi-bias Confidence Regularization \cite{wu-etal-2020-improving}
                     &  & Adversarial SQuAD  \\

Underparameterized Weak Models \cite{sanh-etal-2021-learning}    
                     &                              & Adversarial SQuAD  \\ \midrule

Rationalization \cite{chen-etal-2022-rationalization}     
                     &             \multirow{2}{*}{Intermediate Tasks}                 & Custom  \\

Reasoning Steps \cite{ho-etal-2023-analyzing}     
                     &                             & Custom  \\ \midrule

InfoBERT \cite{wang2021infobert}             
                     & Regularization               & Adversarial SQuAD  \\ \bottomrule
\end{tabular}
}
\caption{Major shortcut mitigation works in MRC and their corresponding evaluation (challenge) sets.
% Major shortcut mitigation works in MRC, along with their type and their corresponding evaluation (challenge) dataset.
}

\label{tab:mitigation-methods}
\end{table*}

\subsection{Summary \& Discussion}
We have covered methods of varying kinds to mitigate or reduce the effect of shortcuts in our models. A table with the list of studies and the challenge set used can be found in Table~\ref{tab:mitigation-methods}.
Adversarial data and bias labels with artifact models are more difficult to apply, since they require a deeper knowledge about the kinds of shortcuts that are present, or involve expensive annotation and verification work. The other methods, including bias labels with weak models, can be applied with more ease.

SQuAD receives the most attention, as most papers showcase their mitigation techniques against Adversarial SQuAD. We highlight the lack of challenge sets for more MRC datasets in Section~\ref{sec_direction}, and encourage further work on mitigation methods that have not seen an application outside of SQuAD.
% yet.

% \cite{jia-liang-2017-adversarial}

\section{Future Directions}
\label{sec_direction}

\paragraph{Evaluating on More Basic NLP Tasks}
When humans can answer the question correctly, humans also can answer several related basic tests that are required for the QA process correctly.
\citet{ribeiro-etal-2020-beyond} introduce the CheckList -- a list of linguistic skills for model evaluation.
However, their list is not complete (e.g., only in MC questions form), and it is automatically generated without using internal knowledge.
To develop safer and more controllable models, we should carefully design a list of basic NLP tasks for comprehensive evaluation.

\paragraph{Building More Community Challenge Sets}

Adversarial SQuAD has been the most used challenge set for studies proposing mitigation techniques in MRC, as is reflected in Table~\ref{tab:mitigation-methods}. Most others either use a custom dataset, or propose one that has not been widely adopted yet. For example, \citet{lee-etal-2021-robustifying} create a challenge set version of HotpotQA by excluding samples where models ignored evidentiality requirements.
We argue that more challenge sets are needed, especially utilizing other MRC datasets as a base. Adopting a wider range of testbeds will allow future mitigation techniques to be more thoroughly evaluated, and more importantly, compared with each other.
%
% An important concern is that these benchmarks may raise the risk of state-of-the-art chasing. We do not believe that they are a perfect solution. However, carefully crafted datasets that can be reused by the community should help give the shortcut mitigation field the uniformity it needs.

\paragraph{Adopting More Mitigation Methods}
We would like to encourage the community to explore the wider range of mitigation methods available. For example, finding a more valuable data subset based on minority samples, or training a weak model for unknown-bias debiasing.

% \paragraph{Training on Intermediate Tasks}
% Although there are many models~\cite{min-etal-2019-multi,fu-etal-2021-decomposing-complex} that utilized the internal reasoning process to build explainable models, 
% they do not use the gold labels for the intermediate tasks~\cite{min-etal-2019-multi} or control all steps in one model to detect the effects of the intermediate tasks.
% %
% Designing a controlled step-by-step model with all intermediate tasks would be interesting to discover the abilities of the models and check whether they still perform shortcuts.

\paragraph{Connecting Intermediate Reasoning Tasks and Chain-of-Thought}

%With the emergence of large language models (LLMs),~
% \citet{wei2022chain} introduce a technique called chain-of-thought (COT) prompting. 
% By using this technique, they show that asking large language models (LLMs) to generate a series of intermediate reasoning steps can improve the performance of LLMs on complex reasoning tasks. 

\rewrite{
\citet{wei2022chain} introduce chain-of-thought (COT) prompting, demonstrating its ability to enhance the performance of large language models (LLMs) on complex reasoning tasks by generating intermediate reasoning steps.
However, it is still unclear whether COT can improve the robustness of LLMs or not.
We believe that exploring the relationships between COT and reasoning shortcuts in the multi-hop MRC would be interesting (e.g., whether COT can overcome the reasoning shortcut issues?).
}

\paragraph{Checking the Validity and Naturalness of Adversarial Examples}
\rewrite{
\citet{morris-etal-2020-reevaluating} evaluate two type of attacks
% (GeneticAttack \cite{alzantot-etal-2018-generating} and TextFooler \cite{Jin_Jin_Zhou_Szolovits_2020})
%
and show that the obtained adversarial examples ``often do not preserve semantics'' and contain grammatical errors.
% Recently, 
\citet{dyrmishi2023humans} survey 378 people about the perceptibility of the adversarial examples.
They reveal that a large portion of the adversarial examples do not pass human quality standards.
It is noted that these two studies were only conducted for text classification and entailment tasks.
% We argue that carefully checking the validity and naturalness of the adversarial examples in the MRC task would be necessary to test whether the models actually fail on the adversarial examples or if the issue lies with the quality of the adversarial examples.
%
We suggest thoroughly examining the validity and naturalness of adversarial examples in the MRC task to determine if model failures are genuine or if the issue stems from the quality of the adversarial examples.
}

% \paragraph{Finding Why the Models Learn Shortcuts}
% \cite{lai-etal-2021-machine}
% \todo{Write more things}

% \paragraph{Accepting Shortcuts}
% As discussed in \citet{Geirhos2020a}, shortcuts also happen with humans and animals.
% Instead of focusing on preventing shortcuts, a potential future direction for this field could be to accept them as an inevitable characteristic of machine learning datasets.
% In that case, models should be able to develop the capacity to use shortcuts conditionally, and achieve human-like performance on OOD and adversarial data.

\section{Conclusion}
\label{sec_conclusion}
We have covered the shortcut identification and mitigation landscape in MRC. 
The presence of shortcuts can be made clear through a variety of methods, and most researchers are aware of this issue. 
Mitigation methods are varied and have some degree of success, but a lot more work is needed before we can achieve models mostly free of shortcut biases. 
Efforts should be made to improve MRC shortcut debiasing techniques by incorporating those found in other fields such as computer vision and NLI, as well as finding methods with lower human and/or computation costs.

\section*{Limitations}
With the rapid growth of the research community and the limited length of the paper, we cannot guarantee that we cover all existing methods in the `Measuring Shortcuts' and `Mitigating Shortcuts' sections. Instead, we summarize and classify the most prominent studies across different approaches and methodologies for each section.
% we strive to include as many important methods as possible for each section.

% It is noted that
% we do not offer an exhaustive citation list. Rather, we summarize and classify the most prominent studies across different approaches and methodologies.

%
% More comprehensive related studies can be found in our GitHub repository.
% We are open to periodically updating our GitHub repository when we come across suitable papers.
%

We also do not have a section discussing why MRC models learn shortcuts. 
We refer readers to \citet{lai-etal-2021-machine} and \citet{du2023shortcut}. 
In summary, the reasons explaining why MRC models learn shortcuts can come from various factors, such as the size and training objective of the LMs, as well as the existing shortcuts in the training set.

Recently, many research papers related to LLMs \cite{LLMSurvey} are being published every day. Various types of promptings \cite{qiao2023reasoning} are being introduced to leverage the abilities of LLMs. Most of the studies we cover in this survey, which focus on measuring shortcuts and mitigating shortcuts, are related to pre-trained language models rather than LLMs. We will leave the study of reasoning shortcuts and the robustness of LLMs for future work.

% \section*{Ethics Statement}

\bibliography{anthology,custom}
\bibliographystyle{acl_natbib}

\appendix

\section{Appendix: Measuring Shortcuts}
\label{sec_measuring_appendix}

\subsection{Adversarial Data Evaluation}
\label{sec_measuring_appendix_adver}

Table \ref{tab:adversarial_dataset} presents the datasets that we mention in Section~\ref{evaluate-adversarial}.

% Label-Preserved
% Label-Changed
% Mix
% Unanswerable questions

\begin{table*}[htp]
  \centering
    % \resizebox{\textwidth}{!}{%
    \begin{tabular}{p{2.872cm}p{12.3cm}}

     \toprule
     \textbf{Type} & \textbf{Datasets}   \\
    \midrule

    % ok
Label-preserved & Adversarial SQuAD \cite{jia-liang-2017-adversarial}, Add4Docs \& Add8Docs \cite{jiang-bansal-2019-avoiding},
AdvRACE \cite{si-etal-2021-benchmarking}
\\ \midrule

% ok
Label-changed & Contrast Sets~\cite{gardner-etal-2020-evaluating}, SAM~\cite{schlegel-atal-2020-semantics}
\\  \midrule

Mix & Natural-Perturbed-QA \cite{khashabi-etal-2020-bang}
\\  \midrule

Adversarial annotation & AdversarialQA \cite{bartolo-etal-2020-beat}, Dynabench \cite{kiela-etal-2021-dynabench} \\ \midrule

Natural distribution shift & Natural-shift-QA \cite{pmlr-v119-miller20a} \\ \midrule

% {,rajpurkar-etal-2018-know,trivedi-etal-2020-multihop,}
Unanswerable questions & Not-answerable questions \cite{nakanishi-etal-2018-answerable}, SQuADRUn \cite{rajpurkar-etal-2018-know}, MuSiQue \cite{trivedi-etal-2022-musique2}
\\ 

      \bottomrule
    \end{tabular}
  \caption{Existing adversarial datasets and their corresponding types. 
  \textit{Mix} denotes the datasets that include both label-preserved and label-changed samples. 
  \textit{Unanswerable questions} indicate the datasets that contain unanswerable questions.
  It is noted that for the two types, \textit{Label-preserved} and \textit{Label-changed}, each method in Table \ref{tab:measure_adversarial} would create a new dataset. For brevity, we only mention some popular adversarial datasets in these two types.
  }
  \label{tab:adversarial_dataset}
\end{table*}

\subsection{Intermediate Task Evaluation}
\label{sec_measuring_appendix_inter}

Table~\ref{tab:intermediate_task} presents all the studies mentioned in Section \ref{sec_detect_inter}.
% We will add our Github page to replace this figure in the released version of the paper. 

% [scale=0.51]
% \begin{figure*}[htp]
%     \includegraphics[scale=0.504]{images/intermediate_tasks}
%     \caption{All studies and datasets that are mentioned in Section \ref{sec_detect_inter}: Intermediate Task Evaluation.
%     We will replace this figure with our Github page in the final version of the released paper.
%     }
%     \label{fig:intermediate_task}
% \end{figure*}

\begin{table*}[htp]
  \centering
    \resizebox{\textwidth}{!}{%
    \begin{tabular}{p{2.0cm}p{1.4cm}p{1.7cm}p{2cm}p{1cm}p{2.5cm}p{4cm}}
     \toprule

     \textbf{Paper} & \textbf{Form} & \textbf{Purpose} & \textbf{Task} & \textbf{Github} & \textbf{Dataset} & \textbf{Note}     \\
    \midrule

\citet{inoue-etal-2020-r4c} & Triple & Evaluation \& Training  & Derivation generation & \href{https://github.com/naoya-i/r4c}{URL} & R4C  & based on HotpotQA \\\midrule

\citet{ho-etal-2020-constructing} & Triple & Evaluation \& Training  & Evidence generation & \href{https://github.com/Alab-NII/2wikimultihop}{URL} &  2WikiMultiHopQA & \\\midrule

\citet{wolfson-etal-2020-break} &  QDMR & Training &  - & 
\href{https://github.com/tomerwolgithub/Break}{URL}
& Break it down  & based on ten datasets (e.g., HotpotQA \& DROP) \\\midrule

\citet{tang-etal-2021-multi} &Sub-question& Evaluation &QA about sub-questions& 
\href{https://github.com/yxxytang/subqa}{URL}
& 1000 samples & based on HotpotQA \\\midrule

\citet{10.1162/tacl_a_00370} & Sub-question& Evaluation \& Training &QA about sub-questions&
\href{https://github.com/eladsegal/strategyqa}{URL}
& StrategyQA &implicit questions \\ \midrule

\citet{ho-etal-2022-well} &Sub-question& Evaluation \& Training &QA about sub-questions&
\href{https://github.com/Alab-NII/HieraDate}{URL}
& HieraDate & only for comparison about Date information \\ \midrule

\citet{trivedi-etal-2022-musique2} &Sub-question& Evaluation \& Training & QA about sub-questions &
\href{https://github.com/stonybrooknlp/musique}{URL}
& MuSiQue & \\ \midrule

\citet{dalvi-etal-2021-explaining}  & Entailment Tree & Evaluation \& Training &tree generation& 
\href{https://github.com/allenai/entailment_bank/}{URL}
& EntailmentBank & based on ARC and WorldTree V2 \\ \midrule

\citet{ribeiro2023street} & Graph & Evaluation \& Training & Graph generation&
\href{https://github.com/amazon-science/street-reasoning}{URL}
& STREET & based on ARC, SCONE, GSM8K, AQUA-RAT, and AR-LSAT \\ 

      \bottomrule
    \end{tabular}
    }
  \caption{All studies and datasets that are mentioned in Section \ref{sec_detect_inter}: Intermediate Task Evaluation.
  % We will replace this table with our Github page in the final version of the released paper.
  }
  \label{tab:intermediate_task}
\end{table*}

\section{Appendix: Mitigating Shortcuts}
\label{sec_mitigating_appendix}

\subsection{Valuable Subsets}
\label{sec_mitigating_appendix_valuable}

As a key paper in this field, \citet{swayamdipta-etal-2020-dataset} track the training dynamics of models during training to identify subsets with special characteristics. The subset with high loss variance throughout training has the highest value, and removing easy-to-learn samples can have a positive impact on generalization.

Following the tracks of this discovery, \citet{yaghoobzadeh-etal-2021-increasing} find the `minority' subset in the training data, which are the samples that are often forgotten by the model while training. An additional training phase with only these samples has surprisingly beneficial debiasing effects, as tested on several challenge sets. Similar work includes \citet{liu2021just}, where the model is re-trained on those samples that had a low loss during the first training pass. Again, performing additional training steps on these subsets has a positive impact on robustness.

Finally, \citet{le2020adversarial} use the notion of predictability to filter out samples with a negative bias influence. Their adversarial algorithm identifies these samples, and their filtered results show promising debiasing performance.

\end{document}